\documentclass[11pt,a4paper]{article}
\usepackage[hyperref]{emnlp2020}
\usepackage{times}
\usepackage{latexsym}
\usepackage{multirow}
\usepackage{booktabs}
\usepackage{tabularx}
\usepackage{graphics}

\usepackage[multiple]{footmisc}

\usepackage{url}

\usepackage{microtype}

\aclfinalcopy 
\def\aclpaperid{540} 

\title{Multi-XScience: A Large-scale Dataset for Extreme Multi-document Summarization of Scientific Articles}

\author{Yao Lu \\
  Mila\\
  University of Waterloo \\
  {\tt lu.yao@ucl.ac.uk} \\\And
  Yue Dong \\
  Mila / McGill University \\
  \texttt{yue.dong2} \\
  \texttt{@mail.mcgill.ca} \\\And
  Laurent Charlin \\
  Mila / HEC Montr\'eal\\
  Canada CIFAR AI Chair \\
  {\tt lcharlin@gmail.com}}

\date{}

\begin{document}
\maketitle

\begin{abstract}
	Multi-document summarization is a challenging task for which there exists little large-scale datasets. We propose Multi-XScience, a large-scale multi-document summarization dataset created from scientific articles. Multi-XScience introduces a challenging multi-document summarization task: writing the related-work section of a paper based on its abstract and the articles it references. Our work is inspired by extreme summarization, a dataset construction protocol that favours abstractive modeling approaches. Descriptive statistics and empirical results---using several state-of-the-art models trained on the Multi-XScience dataset---reveal that Multi-XScience is well suited for abstractive models.\footnote{Our dataset is available at \url{https://github.com/yaolu/Multi-XScience}}
\end{abstract}

\section{Introduction}

\begin{table}[t!]
	\centering
	\small
	\begin{tabularx}{\columnwidth}{|X|}
		\hline
		\textbf{Source 1 (Abstract of query paper)} \\ \hline
		... we present an approach based on ... \textcolor{red}{lexical databases} and ... Our approach makes use of WordNet synonymy information to .... Incidentally, WordNet based approach performance is comparable with the training approach one.\\
		\hline
		\textbf{Source 2 (cite1 abstract)} \\ \hline
		This paper presents a method for the resolution of lexical ambiguity of nouns ... The method relies on the use of the wide-coverage \textcolor{blue}{noun taxonomy of WordNet and the notion of conceptual distance among concepts} ...\\
		\hline
		\textbf{Source 3 (cite2 abstract)} \\ \hline
		Word groupings useful for language processing tasks are increasingly available ... This paper presents a method for \textcolor{orange}{automatic sense disambiguation of nouns appearing within sets of related nouns} ... Disambiguation is performed with respect to WordNet senses ...\\
		\hline
		\textbf{Source 4 (cite3 abstract)}\\ \hline
		In ... \textcolor{purple}{word sense disambiguation}... integrates \textcolor{purple}{a diverse set of knowledge sources ... including part of speech of neighboring words, morphological form} ...\\
		\hline
		\hline
		\textbf{Summary (Related work of query paper)} \\ \hline
		\textcolor{red}{Lexical databases} have been employed recently in \textcolor{purple}{word sense disambiguation}. For example, \textcolor{blue}{... [cite1] make use of a semantic distance that takes into account structural factors in WordNet ...} Additionally, \textcolor{orange}{[cite2] combines the use of WordNet and a text collection for a definition of a distance for disambiguating noun groupings}. ... [cite3] make use of \textcolor{purple}{several sources of information ... (neighborhood, part of speech, morfological form, etc.}) ...\\
		\hline
	\end{tabularx}
	\caption{\small{An example from our Multi-XScience dataset showing the input documents and the related work of the target paper. Text is colored based on semantic similarity between sources and related work.}}
	\label{tab:example}
\end{table}

Single document summarization is the focus of most current summarization research thanks to the availability of large-scale single-document summarization datasets spanning multiple fields, including news (CNN/DailyMail \citep{hermann2015teaching}, NYT \citep{sandhaus2008new}, Newsroom \citep{grusky2018newsroom}, XSum \citep{narayan2018don}), law (BigPatent \citep{sharma2019bigpatent}), and even science (ArXiv and PubMed \citep{cohan2018discourse}).
These large-scale datasets are a necessity for modern data-hungry neural architectures (e.g. Transformers~\citep{vaswani2017attention}) to shine at the summarization task. The versatility of available data has proven helpful in studying different types of summarization strategies as well as both extractive and abstractive models~\citep{narayan2018don}.

In contrast, research on the task of multi-document summarization (MDS) --- a more general scenario with many downstream applications --- has not progressed as much in part due to the lack of large-scale datasets.
There are only two available large-scale multi-document summarization datasets: Multi-News~\cite{fabbri2019multinews} and WikiSum~\cite{j.2018wikisum}. While large supervised neural network models already dominate the leadboard associated with these datasets, obtaining better models requires domain-specific, high-quality, and large-scale datasets, especially ones for abstractive summarization methods. 

We propose Multi-XScience, a large-scale dataset for multi-document summarization using \textit{scientific articles}. We introduce a challenging multi-document summarization task: \textit{write the related work section of a paper} using its abstract (source 1 in Tab.~\ref{tab:example}) and reference papers (additional sources). 

Multi-XScience is inspired by the XSum dataset and can be seen as a multi-document version of extreme summarization~\cite{narayan2018don-xsum}. Similar to XSum, the ``extremeness'' makes our dataset more amenable to abstractive summarization strategies. Moreover, Table~\ref{table:lead-oracle} shows that Multi-XScience contains fewer positional and extractive biases than previous MDS datasets.
High positional and extractive biases can undesirably enable models to achieve high summarization scores by copying sentences from certain (fixed) positions, e.g. lead sentences in news summarization \citep{grenander2019countering-leadbias,narayan2018don}.  Empirical results show that our dataset is challenging and requires models having high-level of text abstractiveness.

\section{Multi-XScience Dataset}
We now describe the Multi-XScience dataset, including the data sources, data cleaning, and the processing procedures used to construct it. We also report descriptive statistics and an initial analysis which shows it is amenable to abstractive models. 

\subsection{Data Source}
Our dataset is created by combining information from two sources: \url{arXiv.org} and the Microsoft Academic Graph (MAG)~\cite{sinha2015mag}. We first obtain all arXiv papers, and then construct pairs of target summary and multi-reference documents using MAG.\footnote{Our dataset is processed based on the October 2019 dump of MAG and arXiv.}

\subsection{Dataset Creation}

We construct the dataset with care to maximize its usefulness. The construction protocol includes: 1) cleaning the latex source of 1.3 millions arXiv papers, 2) aligning all of these papers and their references in MAG using numerous heuristics, 3) five cleaning iterations of the resulting data records interleaved with rounds of human verification.

Our dataset uses a query document's abstract $Q^a$ and the abstracts of articles it references $R^a_1,\ldots, R^a_n$, where $n$ is the number of reference articles cited by $Q$ in its related-work section. The target is the query document's related-work section segmented into paragraphs $Q_1^{rw},\ldots Q_k^{rw}$, where $k$ is the number of paragraphs in the related-work section of $Q$.  We discuss these choices below. Table \ref{tab:example} contains an example from our dataset.

\textbf{Target summary:} 
$Q_i^{rw}$ is a paragraph in the related-work section of $Q$. We only keep articles with an explicit related-work section as query documents. We made the choice of using paragraphs as targets rather than the whole related-work section for the following two reasons: 1) using the whole related work as targets make the dataset difficult to work on, because current techniques struggle with extremely long input and generation targets;~\footnote{10--20 references as input, 2--4 paragraphs as output} and 2) paragraphs in the related-work section often refer to (very) different research threads that can be divided into independent topics. Segmenting paragraphs creates a dataset with reasonable input/target length suitable for most existing models and common computational resources.

\textbf{Source:} the source in our dataset is a tuple ($Q^a,R^a_1,\ldots, R^a_n$).  We only use the abstract of the query because the introduction section, for example, often overlaps with the related-work section. Using the introduction would then be closer to single-document-summarization. By only using the query abstract $Q^a$ the dataset forces models to focus on leveraging the references. 
Furthermore, we approximate the reference documents using their abstract, as the full text of reference papers is often not available due to copyright restrictions.\footnote{Since our dataset relies on MAG for the reference paper as input, some reference papers are not available on arXiv. Our dataset contains all available paper information, including paper ids and corresponding MAG entry.}

\subsection{Dataset Statistics and Analysis}

\begin{table}[h!]
	\centering
	\resizebox{\columnwidth}{!}{
		\begin{tabular}{ l | c c c c } 
			\toprule
			Dataset & \# train/val/test & doc. len  &  summ. len & \# refs  \\ \toprule
			Multi-XScience & 30,369/5,066/5,093 & 778.08 & 116.44 & 4.42 \\
			Multi-News & 44,972/5,622/5,622 & 2,103.49 & 263.66 & 2.79 \\
			WikiSum & 1, 5m/38k/38k & 36,802.5 & 139.4 & 525\\
			\bottomrule
		\end{tabular}
	}
	\caption{\small{Comparison of large-scale multi-document summarization datasets. We propose Multi-XScience. Average document length (``doc.\ len'') is calculated by  concatenating all input sources (multiple reference documents).}}\label{dataset_size_length}
\end{table}

In Table~\ref{dataset_size_length} we report the descriptive statistics of current large-scale multi-document summarization (MDS) datasets, including Multi-XScience. Compared to Multi-News, Multi-XScience has 60\% more references, making it a better fit for the MDS settings. Despite our dataset being smaller than WikiSum, it is better suited to abstractive summarization as its reference summaries contain more novel n-grams when compared to the source (Table~\ref{table:ngram-coverage}).
A dataset with a higher novel n-grams score has less extractive bias which should result in better abstraction for summarization models~\citep{narayan2018don}. Multi-XScience has one of the highest novel n-grams scores among existing large-scale datasets. This is expected since writing related works requires condensing complicated ideas into short summary paragraphs. The high level of abstractiveness makes our dataset challenging since models cannot simply copy sentences from the reference articles.

\begin{table}[h!]
	\resizebox{\columnwidth}{!}{
		\begin{tabular}{ l | c c c c }
			\toprule
			\multirow{2}{*}{Datasets} & \multicolumn{4}{c}{\% of novel n-grams in target summary}  \\
			& unigrams & bigrams & trigrams & 4-grams\\ \toprule
			CNN-DailyMail & 17.00 & 53.91 & 71.98 & 80.29  \\

			NY Times & 22.64 & 55.59 & 71.93 & 80.16  \\
			XSum & 35.76 & 83.45  & 95.50  & 98.49  \\ \bottomrule
			WikiSum & 18.20 & 51.88 & 69.82 & 78.16   \\ 
			
			Multi-News & 17.76 & 57.10  & 75.71  & 82.30   \\ 
			Multi-XScience & \textbf{42.33} & \textbf{81.75}  & \textbf{94.57}  & \textbf{97.62}   \\ \bottomrule
		\end{tabular}
	}
	\caption{\small{The proportion of novel $n$-grams in the target reference summaries across different summarization datasets. The first and second block compare single-document and multi-document summarization datasets, respectively.}
		\label{table:ngram-coverage}}
\end{table}

\begin{table}[h!]
	\centering
	\resizebox{\columnwidth}{!}{
		\begin{tabular}{ l | c c c | c c c } 
			\toprule
			\multirow{2}{*}{Datasets} &  \multicolumn{3}{c|}{\textsc{lead}} & \multicolumn{3}{c}{\textsc{ext-oracle}} \\
			& R-1 & R-2 & R-L & R-1 & R-2 & R-L\\ \toprule
			CNN-DailyMail  & 39.58 & 17.67 & 36.18 & 54.67 & 30.35 & 50.80 \\
			NY Times &  31.85 & 15.86 & 23.75 & 52.08 & 31.59 & 46.72 \\
			XSum &  16.30 &  1.61 &  11.95  &  29.79  &  8.81  & 22.65  \\ \bottomrule
			WikiSum &  38.22 & 16.85 & 26.89 & 44.40 & 22.59 & 41.28 \\
			Multi-News & 43.08 & 14.27  & 38.97  & 49.06  & 21.54  & 44.27  \\ 
			
			Multi-XScience &  \textbf{27.46} & \textbf{4.57}  & \textbf{18.82}  & \textbf{38.45}  & \textbf{9.93}  & \textbf{27.11}  \\ \bottomrule
		\end{tabular}}
		
		\caption{\small{ROUGE scores for
				the \textsc{lead} and \textsc{ext-oracle} baselines for different summarization datasets.}
			\label{table:lead-oracle}}
	\end{table}
	
	Table \ref{table:lead-oracle} reports the performance of the lead baseline\footnote{The lead baseline selects the first-$K$ sentences from the source document as summary.} and the extractive oracle\footnote{The EXT-oracle summarizes by greedily selecting the sentences that maximize the ROUGE-L F1 scores as described in~\citet{nallapati2017summarunner}.} for several summarization datasets. High ROUGE scores on the lead baseline indicate datasets with strong lead bias, which is typical of news summarization \cite{grenander2019countering-leadbias}. The extractive oracle performance indicates the level of ``extractiveness’' of each dataset. Highly-extractive datasets force abstractive models to copy input sentences to obtain a high summarization performance.  
	Compared to the existing summarization datasets, Multi-XScience imposes much less position bias and requires a higher level of abstractiveness from models.
	Both results consolidate that Multi-XScience requires summarization models to ``understand'' source text (models cannot obtain a high score by learning positional cues) and is suitable for abstractive models (models cannot obtain a high score by copying sentences).

	\subsection{Human Evaluation on Dataset Quality}
	
	Two human judges evaluated the overlap between the sources and the target on 25 pairs randomly selected from the test set.\footnote{We invited two PhD students who have extensive research experiences to conduct the dataset quality assessment on our scientific related-work summarization dataset.} They scored each pair using the scale shown in Table~\ref{tab:dataset_quality_criteria}.
	
	\begin{table}[h]
		\centering
		\small
		\begin{tabular}{c|l}
			\toprule
			Score  & Criteria \\
			\toprule
			4  & 75\% - 100\% facts (perfect coverage)\\
			3  & 50\% -75\% facts (major coverage) \\
			2  & 25\% - 50\% facts (partial coverage)\\
			1  & less than 25\% facts (poor coverage)\\
			\bottomrule
		\end{tabular}
		\caption{\small{Dataset quality evaluation criteria}}
		\label{tab:dataset_quality_criteria}
	\end{table}
	
	The average human-evaluated quality score of Multi-XScience is 2.82$\pm$0.4 (95\% C.I.).
	There is a large overlap between the reference abstracts and the targets' related work based on this score~\footnote{This is expected, as it is standard to discuss the key contribution(s) of a paper in its abstract.} which highlights that the major facts are covered despite using only the abstract.

\section{Experiments \& Results}

We study the performance of multiple state-of-the-art models using the Multi-XScience dataset. Detailed analyses of the generation quality are also provided, including quantitative and qualitative analysis in addition to the abstractiveness study.

\subsection{Models}\label{sub_sec:baselines}
In addition to the \emph{lead baseline} and \emph{extractive oracle}, we also include two commonly used unsupervised extractive summarization models, \emph{LexRank} \citep{erkan2004lexrank} and \emph{TextRank} \citep{mihalcea2004textrank}, as baselines. 

For supervised abstractive models, we test state-of-the-art multi-document summarization models \emph{HiMAP}~\cite{fabbri2019multinews} and \emph{HierSumm}~\cite{liu2019hiersumm}. Both deal with multi-documents using a \emph{fusion} mechanism, which performs the transformation of the documents in the vector space. HiMAP adapts a pointer-generator model~\citep{see2017pointer} with maximal marginal relevance (MMR)~\cite{carbonell1998mmr,lebanoff2018mmr} to compute weights over multi-document inputs. HierSumm~\citep{liu2019hiersumm} uses a passage ranker that selects the most important document as the input to the hierarchical transformer-based generation model.

In addition, we apply existing state-of-the-art single-document summarization models, including \emph{Pointer-Generator}~\cite{see2017pointer}, \emph{BART}~\cite{lewis2019bart} and \emph{BertABS} \citep{liu2019bertabs}, for the task of multi-document summarization by simply concatenating the input references.  
Pointer-Generator incorporates attention over source texts as a copy mechanism to aid the generation.
BART is a sequence-to-sequence model with an encoder that is pre-trained with the denosing auto-encoder objective. BertABS uses a pretrained BERT~\citep{devlin2019bert} as the encoder and trains a randomly initialized transformer decoder for abstractive summarization. We also report the performance of BertABS with an encoder (SciBert) pretrained on scientific articles~\cite{beltagy2019scibert}.

\subsection{Implementation Details}
All the models used in our paper are based on open-source code released by their authors. For all models, we use the default configuration (model size, optimizer learning rate, etc.) from the original implementation. During the decoding process, we use beam search (beam size=4) and tri-gram blocking as is standard for sequence-to-sequence models. We set the minimal generation length to 110 tokens given the dataset statistics. Similar to the CNN/Dailymail dataset, we adopt the anonymized setting of citation symbols for the evaluation. In our dataset, the target related work contains citation reference to specific papers with special symbols (e.g. cite\_2). We replace all of these symbols by a standard symbol (e.g. cite) for evaluation.

\subsection{Result Analysis}\label{sub_sec:results}
\noindent\textbf{Automatic Evaluation} We report ROUGE Scores\footnote{The scores are computed with ROUGE-1.5.5 script with option ``-c 95 -r 1000 -n 2 -a -m''} and percentage of novel n-grams for different models on the Multi-XScience dataset in Tables~\ref{tab:experiment-rouge}  and~\ref{table:experiment-ngram-coverage}. When comparing abstractive models to extractive ones, we first observe that almost all abstractive models outperform the unsupervised extractive models---TextRank and LexRank---by wide margins. In addition, almost all the abstractive models significantly outperform the extractive oracle in terms of R-L. This further shows the suitability of Multi-XScience for abstractive summarization.

To our surprise, Pointer-Generator outperforms self-pretrained abstractive summarization models, such as BART and BertABS. Our analyses (Table \ref{table:experiment-ngram-coverage}) reveal that this model performs highly abstractive summaries on our dataset, indicating that the model chooses to generate rather than copy. BART is highly extractive with the lowest novel n-gram among all approaches.  
This result may be due to the domain shift of the self pre-training datasets (Wikipedia and BookCorpus) since the performance of SciBertAbs is much higher in terms of ROUGE-L. In addition, the large number of parameters in the transformer-based decoders require massive supervised domain-specific training data.

\begin{table}[h]
	\centering
	\resizebox{\columnwidth}{!}{
		\begin{tabular}{ l c c c}
			\toprule
			Models & ROUGE-1 & ROUGE-2 & ROUGE-L \\ \toprule
			\multicolumn{4}{l}{Multi-doc Extractive}\\\midrule
			
			\textsc{lead} & 27.46 & 4.57 & 18.82 \\
			
			\textsc{lexrank} & 30.19 & 5.53 & 26.19 \\
			
			\textsc{textrank} & 31.51 & 5.83 & 26.58 \\
			
			\textsc{ext-oracle} & 38.45 & 9.93 & 27.11 \\
			
			\midrule
			
			\multicolumn{4}{l}{Multi-doc Abstractive (Fusion)}\\\midrule
			\textsc{HierSumm(multi)} & 30.02 & 5.04 & 27.60 \\
			\textsc{HIMAP(multi)} & 31.66 & 5.91 & 28.43 \\ 
			\midrule
			
			\multicolumn{4}{l}{Multi-doc Abstractive (Concat)}\\\midrule
			\textsc{BertABS} & 31.56 & 5.02 & 28.05 \\
			\textsc{BART} & 32.83 & 6.36  & 26.61 \\  
			\textsc{sciBertABS} & 32.12 & 5.59 & 29.01 \\
			\textsc{Pointer-Generator} & \textbf{34.11} & \textbf{6.76} & \textbf{30.63} \\

			\bottomrule 
		\end{tabular}
	}
	\caption{ ROUGE results on Multi-XScience test set.  \label{tab:experiment-rouge}}
\end{table}

\begin{table}[h!]
	\resizebox{\columnwidth}{!}{
		\begin{tabular}{ l | c c c c }
			\toprule
			\multirow{2}{*}{Models} & \multicolumn{4}{c}{\% of novel n-grams in generated summary}  \\
			& unigrams & bigrams & trigrams & 4-grams\\ \toprule
			\textsc{PG (CNNDM)} & 0.07 & 2.24 & 6.03 & 9.72  \\
			\textsc{PG (XSUM)} & 27.40 & 73.33& 90.43& 96.04 \\
			\midrule
			\textsc{PG} & 18.82 & 57.54 & 80.22  & 89.32 \\ 
			\textsc{Hiersumm} & 27.52 & 77.16  & \textbf{95.03}  & \textbf{98.51} \\ 
			\textsc{HiMAP} & 23.13 & 63.58  & 86.50  & 94.15 \\ 
			
			\textsc{BART} & 8.15 & 30.13  & 44.53  & 51.75  \\
			\textsc{BertAbs} & \textbf{34.18} & \textbf{81.99} & 95.70  & 98.64 \\
			\textsc{SciBertAbs} & 46.57 & 89.05  & 97.92  & 99.31   \\ \bottomrule
		\end{tabular}
	}
	\caption{The proportion of novel n-grams in generated summary. PG (CNNDM) and PG (XSUM) denotes the pointer-generator model performance reported by papers~\cite{see2017pointer,narayan2018don-xsum} trained on different datasets. All the remaining results are trained on Multi-XScience dataset.} \label{table:experiment-ngram-coverage}
\end{table}

\noindent\textbf{Human Evaluation} We conduct human evaluation on ext-oracle, HiMAP, and Pointer-Generator, since each outperforms others in their respective section of Table~\ref{tab:experiment-rouge}. For evaluation, we randomly select 25 samples and present the system outputs in randomized order to the human judges. Two human judges are asked to rank system outputs from 1 (worst) to 3 (best). Higher rank score means better generation quality. The average score is 1.54, 2.28 and 2.18 for ext-oracle, HiMAP, and Pointer-Generator, respectively. According to the feedback of human evaluators, the overall writing style of abstractive models are much better than extractive models, which provides further evidence of the abstractive nature of Multi-XScience. 

In addition, we show some generation examples in Table~\ref{tab:generation-example}. Since the extractive oracle is copied from the source text, the writing style fails to resemble the related work despite capturing the correct content. In contrast, all generation models can adhere to the related-work writing style and their summaries also the correct content.

\begin{table}[t!]
	\centering
	\small
	\begin{tabularx}{\columnwidth}{|X|}
		\hline
		\textbf{Groundtruth Related Work} \\ \hline
		a study by @cite attempt to address the uncertainty estimation in the domain of crowd counting. this study proposed a scalable neural network framework with quantification of decomposed uncertainty using a bootstrap ensemble ... the proposed uncertainty quantification method provides additional auxiliary insight to the crowd counting model ...\\
		\hline
		\textbf{Generated Related Work (Oracle)} \\ \hline
		in this work, we focus on uncertainty estimation in the domain of crowd counting. we propose a scalable neural network framework with quantification of decomposed uncertainty using a bootstrap ensemble. we demonstrate that the proposed uncertainty quantification method provides additional insight to the crowd counting problem ...\\
		\hline
		\textbf{Generated Related Work (HiMAP)} \\ \hline
		in @cite, the authors propose a scalable neural network model based on gaussian filter and brute-force nearest neighbor search algorithm. the uncertainty of the uncertainty is used as a density map for the crowd counting problem. the authors of @cite proposed to use the uncertainty quantification to improve the uncertainty ... \\
		\hline
		\textbf{Generated Related Work (Pointer-Generator)}\\ \hline
		our work is also related to the work of @cite, where the authors propose a scalable neural network framework for crowd counting. they propose a method for uncertainty estimation in the context of crowd counting, which can be seen as a generalization of the uncertainty ...\\
		\hline
	\end{tabularx}
	\caption{\small{Generation example of extractive oracle (EXT-ORACLE), HiMAP and Pointer-Generator (PG).}}
	\label{tab:generation-example}
\end{table}

\section{Related Work}

Scientific document summarization is a challenging task.  Multiple models trained on small datasets exist for this task~\cite{hu2014automatic,jaidka2013deconstructing,hoang2010towards}, as there are no available large-scale datasets (before this paper). Attempts at creating scientific summarization datasets have been emerging, but not to the scale required for training neural-based models. For example, CL-Scisumm~\cite{jaidka2016overview} created datasets from the ACL Anthology with 30--50 articles; \citeauthor{yasunaga2019scisummnet} and \citeauthor{aburamulti}\footnote{This is concurrent work.} proposed human-annotated datasets with at most 1,000 article and summary pairs. We believe that the lack of large-scale datasets slowed down development of multi-document summarization methods, and we hope that our proposed dataset will change that.

\section{Extensions of Multi-XScience}

We focus on summarization from the text of multiple documents, but our dataset could also be used for other tasks including: 
\begin{itemize}
	\item Graph-based summarization: Since our dataset is aligned with MAG, we could use its graph information (e.g., the citation graph) in addition to the plain text as input.
	\item Unsupervised in-domain corpus: Scientific-document understanding may benefit from using using related work (in addition to other sources such as non-directly related reference manuals). It is worth exploring how to use unsupervised in-domain corpus (e.g., all papers from N-hop subgraph of MAG) for better performance on downstream tasks. 
\end{itemize}

\section{Conclusion}

The lack of large-scale dataset has slowed the progress of multi-document summarization (MDS) research. We introduce Multi-XScience, a large-scale dataset for MDS using scientific articles. Multi-XScience is better suited to abstractive summarization than previous MDS datasets, since it requires summarization models to exhibit high text understanding and abstraction capabilities. Experimental results show that our dataset is amenable to abstractive summarization models and is challenging for current models.

\section*{Acknowledgments}
This work is supported by the Canadian Institute For Advanced Research (CIFAR) through its AI chair program and an IVADO fundamental research grant. We thank Daniel Tarlow for the original idea that lead to this work and Compute Canada for providing the computational resources.

\bibliography{emnlp2020}
\bibliographystyle{acl_natbib}

\end{document}